\documentclass[conference]{IEEEtran}
\IEEEoverridecommandlockouts
\usepackage{cite}
\usepackage{amsmath,amssymb,amsfonts}
\usepackage{algorithmic}
\usepackage{graphicx}
\usepackage{textcomp}
\usepackage{xcolor}
\usepackage{booktabs}
\usepackage{multirow}
\usepackage{kotex}

\usepackage{booktabs}
\usepackage{multirow}
\usepackage[super]{nth}
\def\BibTeX{{\rm B\kern-.05em{\sc i\kern-.025em b}\kern-.08em
    T\kern-.1667em\lower.7ex\hbox{E}\kern-.125emX}}
\begin{document}

\title{Subject-Independent Brain-Computer Interfaces with Open-Set Subject Recognition\\

\thanks{
20xx IEEE. Personal use of this material is permitted. Permission
from IEEE must be obtained for all other uses, in any current or future media, including reprinting/republishing this material for advertising or promotional purposes, creating new collective works, for resale or redistribution to servers or lists, or reuse of any copyrighted component of this work in other works.

This work was partly supported by Institute of Information and Communications Technology Planning and Evaluation (IITP) grants funded by the Korea government (No. 2017-0-00451, Development of BCI based Brain and Cognitive Computing Technology for Recognizing User’s Intentions using Deep Learning; No. 2019-0-00079, Artiﬁcial Intelligence Graduate School Program (Korea University); No. 2021-0-00866, Development of BMI application technology based on multiple bio-signals for autonomous vehicle drivers).}

}

\author{

\IEEEauthorblockN{Dong-Kyun Han}
\IEEEauthorblockA{\textit{Dept. Brain and Cognitive Engineering} \\
\textit{Korea University}\\
Seoul, Republic of Korea \\
dk\_han@korea.ac.kr}\\
\and
\IEEEauthorblockN{Geun-Deok Jang}
\IEEEauthorblockA{\textit{Dept. Artificial Intelligence} \\
\textit{Korea University}\\
Seoul, Republic of Korea \\
gd\_jang@korea.ac.kr}\\
\and
\IEEEauthorblockN{Dong-Young Kim}
\IEEEauthorblockA{\textit{Dept. Artificial Intelligence} \\
\textit{Korea University}\\
Seoul, Republic of Korea \\
dy\_kim@korea.ac.kr}\\

}

\IEEEoverridecommandlockouts
\IEEEpubid{\makebox[\columnwidth]{978-1-6654-6444-4/23/\$31.00~\copyright 2023 IEEE \hfill} \hspace{\columnsep}\makebox[\columnwidth]{ }}

\maketitle

\begin{abstract}

A brain-computer interface (BCI) can't be effectively used since electroencephalography (EEG) varies between and within subjects.
BCI systems require calibration steps to adjust the model to subject-specific data. 
It is widely acknowledged that this is a major obstacle to the development of BCIs.
To address this issue, previous studies have trained a generalized model by removing the subjects' information.
In contrast, in this work, we introduce a style information encoder as an auxiliary task that classifies various source domains and recognizes open-set domains.
Open-set recognition method was used as an auxiliary task to learn subject-related style information from the source subjects, while at the same time helping the shared feature extractor map features in an unseen target.
This paper compares various OSR methods within an open-set subject recognition (OSSR) framework.
As a result of our experiments, we found that the OSSR auxiliary network that encodes domain information improves generalization performance.
\end{abstract}


\section{INTRODUCTION}

The brain-computer interface (BCI) interprets the intention of the user to communicate with external devices by analyzing brain signals\cite{lebedev2006brain,wolpaw2000brain,kim2016commanding}.
Among the various methods for measuring brain signals\cite{naseer2015fnirs, zhang2017hybrid,thung2018conversion,lee2019connectivity}, a well-established and widely used brain signal is electroencephalography (EEG), which is non-invasive and has a high temporal resolution\cite{chen2016high,jeong2020decoding}. The following paradigms are commonly used for EEG-based BCI: motor imagery (MI)\cite{pfurtscheller2001motor,channel,schirrmeister2017deep,jeong2020Brain}, event-related potential (ERP)\cite{fazel2012p300,won2017motion,blankertz2011single,lee2018high}, and steady-state visual potential (SSVEP)\cite{muller2005steady}.

It is challenging to analyze EEGs because they vary over time and between subjects due to psychological or physiological changes. A disadvantage of this intra-subject variability is that it requires subject-specific calibration each time a new user uses the BCI\cite{suk2014predicting}. In addition to collecting subject-specific data and tuning the model, calibration takes approximately 20-30 minutes\cite{suk2011subject,lotte2010regularizing,lee2015subject}. BCI requires a reduction or elimination of this calibration procedure for practical application.

With the rise of deep learning, from a domain shift perspective, many previous studies have proposed transfer learning-based approaches\cite{kim2019subject, multisubjectEMBC, ensembleCNN,kwon2019subject}. In this context, the problem of training generalized BCI models for an unknown subject, known as subject-independent BCI, falls under the domain generalization (DG) problem. 

Previous subject-independent studies for DG purposes, either explicitly or implicitly, have been proposed in the direction of eliminating or ignoring subject (domain) information. It is possible, however, that in a transfer learning process, data from other subjects may have negative effects. In light of this, it may be beneficial to learn subject-specific features simultaneously, i.e., to give the network the ability to distinguish which individuals a sample belongs to.

In \cite{serkan2022}, this issue was addressed by jointly training an auxiliary network that performs an open-set recognition (OSR) task to learn subject-specific style features and to impart invariance between instances of the same subject. OSR aims to provide a system capable of identifying known and unknown classes for real-world scenarios in which unknown classes might be encountered.  The main objective is to lower both the closed-set classification risk associated with labeled known data as well as the open space risk associated with unknown data at the same time\cite{towardosp}. Despite the fact that existing closed-set methods (i.e. softmax) are good at distinguishing classes, their ability to distinguish between known and unknown classes is limited. Therefore, several methods\cite{cpl, cpn} employ prototype learning as a means of forcing training features to resemble the corresponding prototypes, making the distinction between known and unknown easier.

In this study, we validated the effectiveness of the OSR task for learning subject-specific style features in a prototype-based domain generalization framework based on \cite{serkan2022}. 
Various OSR methods, which we refer to as open-set subject recognition (OSSR), were compared against various combinations of subject pools.

\section{METHODS}

OSSR task utilizes the subject labels to impose cross-instance style (subject-specific information) invariance and to learn subject discriminative features, rather than remove user information. A basic framework configuration consists of the following. 
Style encoder and semantic encoder follow convolutional neural network (CNN).
CNN is employed to extract common features between the encoders. 
This style encoder serves as a tool for supporting the classification of subjects by enabling the feature extractor to recognize them. 
Using the OSR method as an auxiliary task to classify subjects reduces the open space risk of potentially unknown subjects and trains a more generalized model.
We also use OSR methods to classify tasks based on semantic features that are derived from the semantic encoder.
An overview of the OSSR framework is shown in Fig. \ref{fig1:Framework}. 
It should be noted that the OSSR framework's goal is not to classify the subjects accurately, but rather to use knowledge from the subjects in order to assist in task classification.

\subsection{Prototype Learning for EEG Decoding}
In this section, we describe the prototype learning methods we used in our experiments.
The prototype learning used in our experiments was originally proposed for open-world problems and open-set recognition.
In this study, we compare various combinations of OSSR frameworks using the convolutional prototype learning (CPL)-based OSR \cite{cpl} used in \cite{serkan2022} and the following prototype learning-based OSR methodologies reciprocal points learning (RPL)\cite{chen2020learning} and adversarial reciprocal points learning (ARPL)\cite{chen2021Adversarial}.

A prototype is an average or representative example of each class that expresses the characteristics of the entire instance of each class.
In contrast to traditional CNNs, prototype learning does not use a softmax layer but instead learns prototypes based on a data set. Here, prototypes are learnable representations formed by one or more latent features. 
The following sections provide a detailed description of the three methodologies described above.

\subsubsection{Generalized Convolutional Prototype Learning}
Yang\textit{ et al.} \cite{cpl} proposed the Generalized Convolutional Prototype Learning (GCPL). 
In GCPL prototypes are trained in conjunction with a feature extractor and instances are classified according to the most similar prototype. Additionally, a prototype loss (PL) is proposed as a regularization for enhancing the intra-class compactness of the representation.

In \cite{serkan2022}, they used hybrid loss with distance-based cross-entropy loss (DCE) loss and PL, just like GCPL. At this time, the GCPL loss was also used for the semantic encoder ($\mathcal{L}_{clf}$).
Following that, we add an experiment with a framework in which the semantic encoder is trained with cross-entropy loss, i.e., the semantic task is a closed-set classification.

In this approach, the distance between the samples and the prototypes is considered the probability of that sample belonging to the class of that prototype. 
Let's assume there are $C$ classes and each has one prototype.
Considering the training sample $(x,y)$, let $M_{i}$ represents the prototype where $i$ $\in$ ($1,2,...,C$). 
The prototypes $M = \{m_i | i=1,...,C\}$ are learned during training. 

The probability that the sample $x$ belongs to class $k$ is related to the probability that the extracted feature belongs to the prototype $m$.
The probability for the prototype $m$ is measured by the corresponding distance and softmax as: 
\begin{equation}
p(y = k | x, M) = \frac{e^{-\gamma(\parallel f(x) - m_i \parallel_2^2)}}{\sum_{k=1}^C e^{-\gamma(\parallel f(x) - m_k \parallel_2^2)}}
,\end{equation}
where $f(x)$ is the CNN-based feature extractor and $\gamma$ is a temperature parameter that controls the probability assignment hardness. DCE is the cross-entropy loss calculated using this probability. 
As a regularizer, PL is added as follows:
\begin{equation}
l_{p} = \parallel f(x) - m_i \parallel_2^2
 .\end{equation}
As a result, the combined loss function is as follows:
\begin{equation}
l_{gcpl} = l_{dce} + \beta l_{p}.
\end{equation}
The weight of the PL is controlled by $\beta$.

\begin{figure}[t!]
    \centering
    \includegraphics[width=0.9\linewidth]{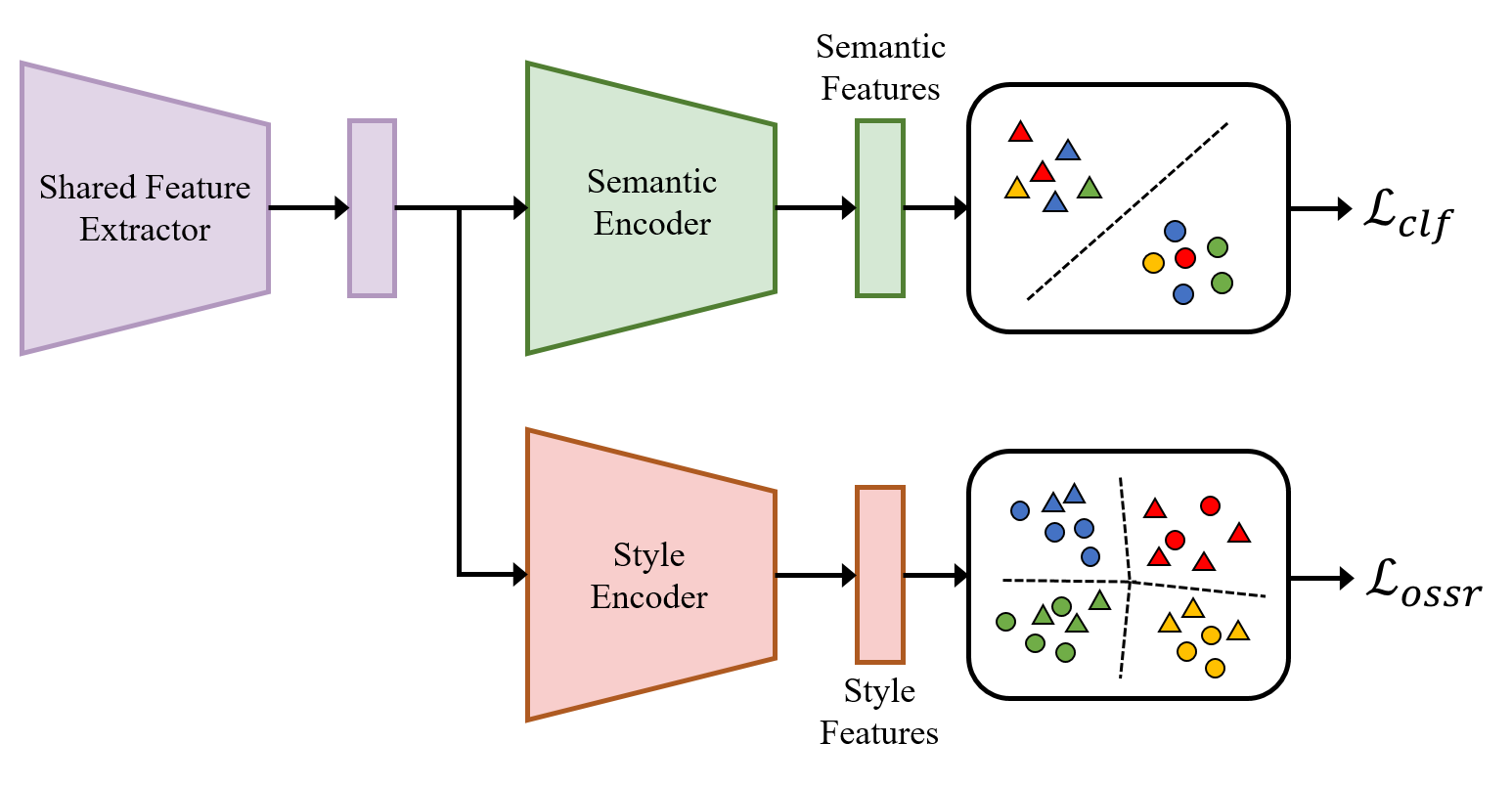}
    \caption{The overview of the proposed framework. Our framework consists of two encoders, one for style and one for semantic. In feature space, different colors denote different subjects, and different shapes denote different classes.}
    \label{fig1:Framework}
\end{figure}

\subsubsection{Reciprocal Points Learning}
The RPL\cite{chen2020learning} introduced a new concept, Reciprocal Point, which represents the extra-class space corresponding to each known class.
As with GCPL, the learnable parameter reciprocal point $M$ is calculated by using the following loss function.
\begin{equation}
p(y = k | x, M) = \frac{e^{\gamma(\parallel f(x) - m_i \parallel_2^2)}}{\sum_{k=1}^C e^{\gamma(\parallel f(x) - m_k \parallel_2^2)}},
\end{equation}
\begin{equation}
l_{rp} = -\log p(y=k | x, M)
 .\end{equation}

RPL introduced a regularization term that limits the distance between the closed-set sample and reciprocal point to some extent. 
\begin{equation}
l_o(x;M^k,R^k) = MSE(\parallel f(x) - m_k \parallel_2^2, R^k)
 ,\end{equation}
where $R^k$ is the radius (learnable margin) initialized to 1.
The combined loss function is as follows:
\begin{equation}
l_{rpl} = l_{rp} + \gamma l_{o},
\end{equation}
where $\gamma$ is a hyperparameter to control the weight of reducing open space
risk module.

\subsubsection{Adversarial Reciprocal Points Learning}

ARPL\cite{chen2021Adversarial} is based on RPL, but the distance and constraint functions have been changed.
The distance $d(f(x),M^k)$ between $x$ and reciprocal point $M^k$ can be obtained by combining the Euclidean distance $d_e$ with the dot product $d_d$:
\begin{equation}
d(f(x),M^k) = d_e(f(x),M^k) - d_d(f(x),M^k).
\end{equation}
The classification probability can be expressed as follows:
\begin{equation}
p(y = k | x, M) = \frac{e^{\gamma d(f(x),M^i)}}{\sum_{k=1}^C e^{\gamma d(f(x),M^k)}}.
\end{equation}
In ARPL, the regularization term was changed as follows:
\begin{equation}
l_o(x;M^k,R^k) = MAX(d_e(f(x),M^k)-R,0)
.\end{equation}
Margin $R$ is also initialized to 1.

\subsection{Open-Set Subject Recognition framework}
As shown in Fig. \ref{fig1:Framework}, in order to separate subject information from class information, we employ two encoders referred to as style and semantic encoders. In our framework, each encoder can be trained by various types of Loss.
We train the style and semantic encoder using the hybrid loss function as follows:
\begin{equation}
\mathcal{L} = \mathcal{L}_{clf} + \alpha \mathcal{L}_{ossr},
\end{equation}
where $\mathcal{L}_{clf}$ means the loss function calculated for semantic encoder, and $\mathcal{L}_{ossr}$ means the loss function calculated for style encoder. $\alpha$ controls the weight of the OSSR task.

\section{EXPERIMENTS}

\subsection{Dataset and Data-Split} 

OpenBMI dataset \cite{openbmi}: 
The dataset includes 54 subjects and two motor imagery classes (left hand, right hand). 
Data for each subject consists of 4 sessions: offline and online sessions for two days.
There are 100 trials in each session.
Each trial consists of 4 seconds of EEG recorded at 62 channels, 1000Hz, and we downsampled to 250Hz for this experiment.
We used Leave-one-subject-out cross-validation.
Data from the source dataset was divided into two parts, 8:2, and used as training and validation data, respectively.
For evaluation, only the fourth session of the test subject was considered\cite{adaptivenn}.
We experimented on datasets consisting of four different sizes of subject numbers.
There were 11 subjects, 21 subjects, 31 subjects, and 54 subjects. 
The runs contain different sets of subjects. However, for each method, the same sets are used. Five runs are conducted with 10 subjects, three runs with 20 subjects, two runs with 30 subjects, and one run with 53 subjects.
The procedure of all training follows the data split configuration of \cite{serkan2022}.

\begin{table}[t]
\caption{Motor imagery classification performance (Accuracy (\%)). Results are averaged among runs.}
\label{tab:results}
\centering
\resizebox{\columnwidth}{!}{%
\begin{tabular}{@{}lcccc@{}}
\toprule
\multicolumn{1}{c}{\multirow{2}{*}{Method}} & \multicolumn{4}{c}{\#   subject}                          \\ \cmidrule(l){2-5} 
                        & 10           & 20           & 30           & 53           \\ \midrule
Baseline \cite{serkan2022}                       & 72.83 ($\pm$14.22) & 80.65 ($\pm$13.02) & 82.81 ($\pm$12.83) & 84.98 ($\pm$12.18) \\
GCPL$_{clf}$\cite{serkan2022} & 72.67 ($\pm$14.04) & 80.30 ($\pm$12.45) & 82.22 ($\pm$12.11) & 84.80 ($\pm$11.84)  \\
GCPL$_{clf}$+GCPL$_{ossr}$  \cite{serkan2022} & \textbf{74.17} ($\pm$14.19) & \textbf{81.61} ($\pm$12.83) & \textbf{84.07} ($\pm$12.25) & 85.22 ($\pm$12.24) \\ 
CE$_{clf}$+GCPL$_{ossr}$  & 73.67 ($\pm$12.78) & 79.99 ($\pm$12.88) & 82.97 ($\pm$12.11) & 84.48 ($\pm$12.41) \\
RPL$_{clf}$+RPL$_{ossr}$  & 70.00 ($\pm$12.64) & 77.39 ($\pm$13.99) & 81.81 ($\pm$12.89) & 84.06 ($\pm$12.44) \\
CE$_{clf}$+RPL$_{ossr}$  & 71.96 ($\pm$12.99) & 79.73 ($\pm$13.19) & 82.86 ($\pm$12.11) & 84.57 ($\pm$12.17) \\
ARPL$_{clf}$+ARPL$_{ossr}$  & 73.34 ($\pm$13.73) & 80.35 ($\pm$13.64) & 82.39 ($\pm$12.89) & \textbf{85.33 }($\pm$11.73) \\
CE$_{clf}$+ARPL$_{ossr}$  & 72.71 ($\pm$12.71) & 80.08 ($\pm$13.34) & 82.96 ($\pm$12.16) & 84.72 ($\pm$11.71) \\ \bottomrule
\end{tabular}%
}
\end{table}

\subsection{Experimental Details}
Following previous studies \cite{adaptivenn, serkan2022}, DeepConvNet \cite{schirrmeister2017deep} was used as a shared backbone feature extractor. In both style and semantic encoders, a fully connected layer (1400 $\times$ 2) was used.
We set the case where $\mathcal{L}_{clf}$ is cross-entropy loss and $\mathcal{L}_{ossr}$ is not used as the baseline, and the case where $l_{gcpl}$ is used for both $\mathcal{L}_{clf}$ and $\mathcal{L}_{ossr}$ is marked as \cite{serkan2022}.
$\beta$ was set to 0.001 and the weight of $\mathcal{L}_{ossr}$ ($\alpha$) was set to 0.1.
We experimented with cases where cross-entropy or $l_{rpl}$ or $l_{arpl}$ was used for $\mathcal{L}_{clf}$ and $l_{rpl}$ or $l_{arpl}$ was used for $\mathcal{L}_{ossr}$. 
When $l_{rpl}$ was used,$\gamma$ was set to 0.001.
In all cases, The weight of $\mathcal{L}_{ossr}$ ($\alpha$) was set to 0.1.
Adam was used as the optimizer and trained at a learning rate of 0.005, and a cosine annealing learning rate scheduler was also used.

\section{RESULTS AND DISCUSSION}

Various OSSR results were compared with existing subject-independent methods. 
In total, we experimented with five different methods in subject-independent settings. 
The first is that the semantic encoder is trained with CE, while a style encoder is trained with GCPL and the second is the RPL is used for both the semantic encoder and style encoder. 
Thirdly, semantic encoders use CE, and style encoders use RPL.
Fourth, ARPL is applied to both semantic and style encoders.
Lastly, CE is applied to semantic encoders while ARPL is applied to style encoders.
Comparing the results can be found in Table \ref{tab:results}.

CE$_{clf}$+ GCPL$_{ossr}$ achieves $73.67\%$ average accuracy on 10 subjects.
RPL$_{clf}$+ RPL$_{ossr}$ and CE$_{clf}$+ RPL$_{ossr}$ showed poor performance in 10 subjects, but good performance compared to the baseline in 30 and 53 and showed that reciprocal points can be used as a classifier.
ARPL$_{clf}$+ ARPL$_{ossr}$ and CE$_{clf}$+ ARPL$_{ossr}$ showed similar performance to the baseline in all cases and was higher than RPLs in most cases.
To summarize, ARPL$_{clf}$+ARPL$_{ossr}$ achieved the highest performance in 53 subjects, and GCPL$_{clf}$+ GCPL$_{ossr}$ showed the highest performance in the other cases. However, the proposed OSSR framework showed good performance compared to the baseline in the settings of 10 and 30 subjects and showed a slight difference even with 20 subjects.
As mentioned in \cite{serkan2022}, performance increased as the number of subjects increased, but the gap between methods narrowed.

\section{CONCLUSION}
In this paper, we proposed an open-set subject recognition framework for subject-independent BCIs.
In our framework, open-set recognition was used as an auxiliary task to encode subjects' information. 
The proposed framework has, to the best of our knowledge, achieved the best performance on the OpenBMI dataset using subject-independent settings. 
In the future, we intend to introduce a framework for subject-adaptive classification (domain adaptation). We also plan to apply the OSR framework, which classifies previously unseen classes, to real-world BCIs.

\bibliographystyle{IEEEtran}
\bibliography{REFERENCE}

\end{document}